\documentclass[runningheads]{llncs}

\usepackage{eccv}

\usepackage{eccvabbrv}

\usepackage{etoolbox}
\AtBeginDocument{%
  \setlength{\abovedisplayskip}{6pt}%
  \setlength{\belowdisplayskip}{6pt}%
  \setlength{\abovedisplayshortskip}{6pt}%
  \setlength{\belowdisplayshortskip}{6pt}%
}

\usepackage{graphicx}
\usepackage{booktabs}
\usepackage{adjustbox}
\usepackage{multirow}
\usepackage{subcaption}
\usepackage{threeparttable}

\usepackage{amsfonts}       
\usepackage{nicefrac}       
\usepackage{microtype}      
\usepackage[bottom]{footmisc}
\usepackage[dvipsnames]{xcolor}
\usepackage{tabularx}

\usepackage{etoolbox}
\preto\equation{\vspace{-0.15em}}
\appto\endequation{\addvspace{-0.15em}}

\newcommand{\mapehead}{%
  \shortstack[c]{MAPE\\[-1pt]{\scriptsize(\%)}}%
}

\usepackage[accsupp]{axessibility}  

\usepackage{hyperref}

\usepackage{orcidlink}
\usepackage{amsmath}

\newcommand{\projname}{EgoHRV\xspace}

\begin{document}

\title{EgoHRV: Continuous Heart Rate Variability Estimation from Egocentric Systems for \\ Autonomic Response and Skill Assessment}

\titlerunning{EgoHRV: HRV Estimation for Autonomic Response and Skill Assessment}

\author{Berken Utku Demirel\orcidlink{0000-0002-9026-3693} \and
Christian Holz\orcidlink{0000-0001-9655-9519}}

\authorrunning{B. U. Demirel and C. Holz}

\institute{Department of Computer Science, ETH Zürich, Zürich, Switzerland\\[1em]
\small\href{https://siplab.org/projects/EgoHRV}{\color{magenta}{\texttt{https://siplab.org/projects/EgoHRV}}}}

\maketitle

\begin{figure}[t]
  \centering
  \includegraphics[width=\textwidth]{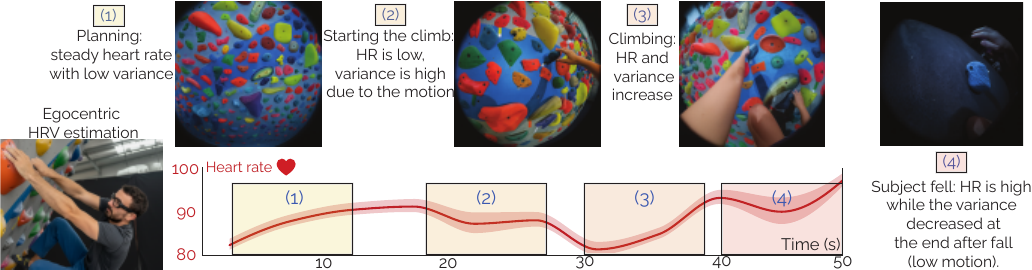}
  \caption{\projname continuously estimates heart rate as well as heart rate variability with uncertainty from gaze cameras in egocentric systems to enrich understanding of human behavior, cognitive effort, stress, and skill assessment.}
  \label{fig:teaser}
\end{figure}

\begin{abstract}
Egocentric vision systems capture human behavior from visible cues, but overlook physiological indicators of autonomic states such as stress, engagement, and attention.
Heart rate variability (HRV) is a widely used noninvasive marker of autonomic regulation under stress.
HRV reflects small timing differences between successive heartbeats and has so far been out of reach for egocentric platforms, where motion and noise in gaze video mask exactly this fine-grained timing.
We propose \emph{EgoHRV}, a method that estimates HRV as well as heart rate (HR) from the gaze cameras that are already integrated into egocentric headsets.
Our pipeline combines a 3D backbone with a novel low--high decomposition module that extracts the blood volume pulse (BVP) signal from gaze video.
Our cross-domain pretraining aligns the frequency-domain representations of contact-based and camera-derived signals.
This alignment gives EgoHRV the temporal precision to recover HRV from the subtle fluctuations in gaze video.
EgoHRV achieves state-of-the-art accuracy for HR and HRV estimation from egocentric video, and its uncertainty-aware design improves downstream behavioral modeling.
Integrating our HRV estimates and confidence measures into EgoExo4D's proficiency estimator raises accuracy by 17.8\%.
Beyond skill, continuous HRV estimation also opens egocentric systems to stress- and arousal-aware estimation tasks.
\\[.2em]
Code: \url{https://github.com/eth-siplab/EgoHRV}
\vspace{-1mm}
\end{abstract}


\vspace{-5mm}
\section{Introduction}
\label{sec:intro}
Egocentric vision systems, such as Magic Leap~\cite{noauthor_magic_nodate} or Project Aria~\cite{engel2023projectarianewtool}, can aid in understanding human behavior and interaction from a first-person perspective~\cite{ego4D}.
These wearable devices capture both the surrounding environment and the wearer's actions by providing rich multimodal data~\cite{ego4D, ego_exo_4d} for studying activity and social interaction~\cite{sigurdsson2018actor}.
Recent work has analyzed egocentric video to reveal behavioral patterns in daily tasks that static third-person systems often miss~\cite{Damen2018EPICKITCHENS, Damen2021PAMI, ego_exo_4d}.
As their sensing capabilities expand to eye tracking and inertial cues, egocentric platforms can now enable research on context-aware behavior modeling and human sensing more broadly~\cite{old_ego, egolife, EgoEnv}.

Physiological indicators play a central role in modeling human behavior~\cite{emotion_recognition, IEEE_emotion, jammot2025egoemotion}.
While a person's motion or gaze can be directly observed by vision models~\cite{perceptual_integration}, physiological signals are subtle or invisible, yet they reveal latent human autonomic~\cite{dawn_of_sensors} and affective~\cite{Pruski10} processes.
A common measure of physiological state is heart rate (HR), which reflects overall autonomic activity~\cite{HRV_cardiac_autonomic_system} and offers insight into how people respond to tasks~\cite{LUQUECASADO201683}, environments~\cite{environments}, and social contexts~\cite{Cacioppo1990}.
While recent work showed that HR can be recovered from eye-gaze cameras on egocentric systems (egoPPG~\cite{egoPPG}), estimates have coarse temporal resolution (60\,s averages) and cannot capture a person's short-term responses.

More than mere heart \emph{rate},
heart rate \emph{variability} (HRV) captures autonomic regulation and is thus a rich noninvasive marker of stress~\cite{hrv_stress, rajendra_acharya_heart_2006, kim2018stress, thayer2012metaanalysis, castaldo2015acute}, cognitive effort~\cite{cognitive}, and emotional reactivity~\cite{hypertension_HRV, HRV_mental_disorders}.
Unlike HR, HRV captures the relation between sympathetic and parasympathetic branches of the autonomic nervous system~\cite{hypertension_HRV, HRV_cardiac_autonomic_system, taskforce1996hrv}.
This modulation thus helps link moment-to-moment physiological changes to perception, decision-making, and human behavior~\cite{elite_performers, decision_making_hrv, HRV_emotional_events}.
HRV also reveals mental workload during complex tasks, attention in learning or driving, and stress adaptation in natural environments~\cite{hrv_mental_workload, HRV_mentalWorkload, hrv_driving}.
In situated interactions, moments of fear, frustration, time pressure, and safety-critical decisions produce autonomic responses that manifest in HRV decreases alongside increased electrodermal activity~\cite{luong2022physiological, giannakakis2022review, sharma2012objective}.
Continuous tracking of HRV on egocentric systems can thus moves research in this domain beyond the current focus on action and activity tracking, because the same signals capture indicators of stress, engagement, or fatigue in real time and supports downstream tasks such as proficiency assessment and stress-aware interaction.

%

However, accurately decoding HRV from video is challenging because it requires beat timing at a millisecond level, as even small timing errors substantially distort HRV estimates.
This makes it highly sensitive to motion and noise common in mobile scenarios, in which egocentric systems are used.
Additional factors such as head movement and posture changes further degrade signal quality~\cite{DeepPPG, HRV_artifacts}.
As we show in this paper, methods designed for HR estimation from video fail to generalize to estimating HRV because of the key requirement to preserve temporal structure rather than just the overall trend~\cite{HRV_fingertip, BCG_JBHI}.

In this paper, we introduce a method for estimating continuous HR and HRV as a basis for stress- and skill-aware modeling in egocentric systems.
We demonstrate its benefits for behavioral understanding in egocentric vision through the proficiency estimation benchmark task.
Since HRV is an established noninvasive proxy for autonomic stress responses~\cite{thayer2012metaanalysis, kim2018stress}, our continuous estimates can support stress- and arousal-aware tasks on egocentric systems without additional body-worn sensors.
Taken together, we contribute:
\begin{enumerate}
    \item an uncertainty-aware HR \& HRV estimator from egocentric gaze video via a 3D backbone with low–high decomposition, setting state-of-the-art results,
    \item a large-scale cross-domain pretraining strategy that leverages contact-based blood volume pulse (BVP) datasets with ground truth and uses frequency-domain alignment to transfer physiologically meaningful representations to our egocentric vision approach, and
    \item a multimodal PhysFusion module that fuses HRV with video representations to improve proficiency estimation on EgoExo4D by 17.8\% and supply a continuous HRV signal to downstream stress- and arousal-aware applications.
\end{enumerate}



\section{Related Work}
\label{sec:related_work}

\paragraph{Egocentric vision.}
Egocentric vision research has rapidly expanded with the advent of wearable mixed-reality devices, e.g., Magic Leap~\cite{noauthor_magic_nodate} and Project Aria~\cite{engel2023projectarianewtool}. 
These platforms enable continuous first-person perception for understanding user actions, surroundings, and interactions~\cite{EgoChoir, ego_exo_4d}.
Much of the work related to these platforms has focused on visual understanding tasks, including action recognition~\cite{ego_action_recognition, action_recognition, ego_action_IPL}, anticipation~\cite{ego_anticipation, furnari2018Leveraging}, and hand–object interaction~\cite{EgoChoir, hoi4}, as well as 3D reconstruction, and mapping~\cite{xu2024macego3d, gu2024egolifter}.

Large-scale datasets, Ego4D~\cite{ego4D} and EgoExo4D~\cite{ego_exo_4d}, have accelerated the research by providing thousands of hours of first-person recordings across diverse contexts and users.
These datasets enabled progress on social interaction modeling and multimodal perception tasks that integrate gaze, speech, and motion~\cite{human_gaze_activity_recognition}.
However, even with these developments, physiological sensing and autonomic state estimation remain absent from existing egocentric benchmarks, leaving an opportunity to connect external behavior with internal physiological state.

\vspace{3mm}
\noindent
\textbf{Physiological sensing in vision.}
Wearable sensors have enabled continuous monitoring of cardiovascular and affective states in daily life~\cite{PPG_analysis, DeepPPG, moebus2024nightbeat}.
In parallel, camera-based methods estimate BVP from subtle skin-color changes, commonly referred to as remote photoplethysmography (rPPG), caused by cardiac activity~\cite{early_rPPG}.
Traditional rPPG methods rely on chrominance or blind-source separation techniques~\cite{CHROM}, while more recent approaches adopt spatiotemporal networks~\cite{physformer, FactorizePhys, yu2019remote} to improve robustness under varying illumination and skin tone.
Beyond heart rate, camera-based approaches also decode further autonomic indicators, such as sympathetic arousal, which can be recovered from peripheral blood-flow changes in video~\cite{braun2023videobasedsympathetic, braun2024sympcam}.

Despite much progress, most rPPG datasets and models assume fixed cameras and stationary subjects~\cite{Niu2018VIPLHRAM, MMSE_HR, OBF_dataset, braun2024sympcam}, which limits their applicability to wearable scenarios.
In egocentric systems, cameras are mounted on the head and undergo significant motion, making direct transfer of conventional rPPG methods unreliable.
Recent work, egoPPG~\cite{egoPPG}, addressed this limitation by estimating HR from eye-tracking videos and inertial cues.
Although egoPPG demonstrated the feasibility of extracting coarse HR trends, its estimates averaged over long windows and achieved errors around 8--9\,bpm, which is far below the temporal fidelity and accuracy of contact sensors~\cite{beliefppg}.

\vspace{3mm}
\noindent
\textbf{Heart rate variability estimation.}
HRV provides a more sensitive measure of autonomic function than HR by capturing short-term fluctuations in cardiac rhythm~\cite{hypertension_HRV}.
HRV estimation from non-contact sensors remains a major challenge.
Accurate HRV computation requires millisecond-level timing precision in inter-beat intervals, which is difficult to achieve from video affected by noise, motion, and low sampling rates~\cite{HRV_artifacts, HRV_fingertip, BCG_JBHI}.
In egocentric systems, additional challenges arise from variable head motion, changing illumination, and limited skin visibility, all of which disrupt consistent waveform morphology.
PulseFormer, an egocentric method for HR, estimates a single HR value per 60\,s~\cite{egoPPG}, discarding the dynamics that characterize real-world behavior.
In contrast, \projname estimates short-term mean inter-beat intervals and derives HRV at this higher temporal resolution.

\vspace{3mm}
\noindent
\textbf{HRV as a proxy for perceived stress.}
Stress activates the autonomic nervous system, which modulates the cardiac dynamics that HRV indexes~\cite{chrousos2009stress}.
HRV widely serves as an established peripheral indicator of adaptive responses to challenge~\cite{thayerLane2000neurovisceral, thayer2012metaanalysis} and is a standard input feature for biosignal-based stress classification~\cite{sharma2012objective, giannakakis2022review}.
It drops under perceived stress~\cite{kim2018stress, castaldo2015acute}, including in situated and immersive settings~\cite{luong2022physiological}.
HRV thus serves as a stress-related autonomic marker rather than a direct sympathetic readout~\cite{taskforce1996hrv, shaffer2017overview, laborde2017hrv}.


\section{Problem: HRV is a highly sensitive metric}
\label{sec:motivation}
HRV estimation from egocentric video is challenging due to algorithmic and data limitations.  
We highlight two key issues that motivate our approach.  


\subsection{Error Accumulation}
Existing approaches for physiological estimation from vision, including models for rPPG and egoPPG tasks~\cite{egoPPG, yu2019remote, FactorizePhys, liu2022rppg, luo2024physmamba, zou2024rhythmmamba}, reconstruct the BVP indirectly by predicting frame-to-frame differences of skin reflectance and regressing them toward the temporal derivatives of the reference PPG signal.  
During inference, these predicted differences are cumulatively summed to recover the waveform.  
While this formulation improves stability and reduces spurious correlations with the background, it introduces a drift.  
Small frame-wise prediction errors accumulate over time, leading to deviations in the integrated signal.  
This issue is further amplified in egocentric settings, where head motion and dynamic illumination can cause larger variations than in stationary rPPG.  
As a result, even minor prediction noise can distort the reconstructed waveform.  
Such cumulative bias is detrimental for HRV estimation, which depends on millisecond-level accuracy rather than waveform trends.

Let $\hat{d}_t$ denote the model’s predicted BVP difference at frame $t$, and $d_t$ the ground-truth derivative.
The reconstructed BVP is obtained by integration.
\begin{equation}
    \hat{p}_t = \sum_{i=1}^{t} \hat{d}_i, \quad p_t = \sum_{i=1}^{t} d_i.
\end{equation}
The instantaneous reconstruction error is given in Equation~\ref{eq:inst_reconst}.
\begin{equation}\label{eq:inst_reconst}
    \epsilon_t = \hat{p}_t - p_t = \sum_{i=1}^{t} (\hat{d}_i - d_i),
\end{equation}
and even if the per-frame prediction noise is zero-mean with variance $\sigma^2$, its cumulative variance increases with time:
\begin{equation}
    \mathrm{Var}[\epsilon_t] = t\,\sigma^2.
\end{equation}
As $t$ grows, small local prediction errors compound, progressively distorting waveform morphology.
This drift propagates directly into inter-beat intervals, producing large HRV estimation errors even when frame-wise losses remain small.
Given HRV’s extreme sensitivity to timing, derivative-based reconstruction schemes become unreliable for dynamic, motion-heavy egocentric data.

\subsection{Limited use of large-scale physiological datasets}
Although egocentric datasets with synchronized ground-truth physiological recordings are limited, large-scale public BVP datasets already exist from mobile settings~\cite{wildppg, DeepPPG}.
These datasets capture a wide range of heart rate and waveform under varying illumination, providing valuable priors.
Yet, current methods have not leveraged these broader data sources for pretraining.
Learning BVP representations from such datasets and then adapting them for visual inputs from cameras can significantly improve the performance.
%

\section{Method}
\label{sec:method}
Our method has two aims.
First, we estimate HR and HRV directly from egocentric video.
Second, we leverage these physiological estimates to improve downstream proficiency prediction in large-scale egocentric datasets.  
The following sections describe each component in detail.

\subsection{HR \& HRV Estimation}
HR and HRV are both determined by the sequence of inter-beat intervals (IBIs).
HR is the reciprocal of the mean IBI over a window, while HRV is the variability of consecutive IBIs over the same window.
Estimating IBIs once therefore gives both metrics without separate heads or losses.
We design our pipeline to (a) recover a clean BVP waveform from video (Section~\ref{sec:3d_backbone}) and (b) predict IBIs from that waveform using cross-domain supervision (Section~\ref{sec:crossdomain}); HR and HRV are then computed from the predicted IBIs.

\subsubsection{3D backbone with low-high decomposition}
\label{sec:3d_backbone}
We adopt a 3D convolutional architecture~\cite{egoPPG,yu2019remote} that processes temporal segments of $T{=}128$ frames (4.3\,s) at a spatial resolution of $48{\times}128$ pixels. 
Each input clip $\mathbf{x}$ and its output representation $\mathbf{y}$ are defined as in Equation~\ref{eq:inp_out}.
\begin{equation}\label{eq:inp_out}
\mathbf{x} \in \mathbb{R}^{T\times H\times W}, 
\qquad 
\mathbf{y} = f_{\text{3D}}(\mathbf{x}) \in \mathbb{R}^{1\times T},
\end{equation}
where $(H,W)$ denote the spatial dimensions, and $T$ is the time (frames).  
The representation $\mathbf{y}$ captures frame-wise reflectance differences that encode temporal variations in skin appearance.
Our 3D backbone model predicts temporal differences rather than absolute intensity to emphasize dynamic changes similar to previous work~\cite{egoPPG, FactorizePhys}.

However, raw differencing behaves as a high-pass filter, amplifying motion noise.  
We aim to mitigate this effect with a learnable low–high decomposition that explicitly separates high frequency noise from pulse.
We define the low-frequency component and the complementary high-frequency component as in Equation~\ref{eq:high_low}.
\begin{equation}\label{eq:high_low}
\mathbf y_{\mathrm{lo}}=\mathbf y * h,\hspace{3mm}
h=\tfrac{1}{k}\,\mathbf 1_k,\hspace{1mm} k{=}11,\hspace{3mm}
\mathbf y_{\mathrm{hi}}=\mathbf y-\mathbf y_{\mathrm{lo}},
\end{equation}
Two lightweight convolutional heads refine these signals:
\begin{equation}
\tilde{\mathbf{y}}_{\mathrm{lo}} = f_{\mathrm{lo}}(\mathbf{y}_{\mathrm{lo}}), 
\qquad 
\tilde{\mathbf{y}}_{\mathrm{hi}} = f_{\mathrm{hi}}(\mathbf{y}_{\mathrm{hi}}),
\end{equation}
and their outputs are fused through a gated residual connection with a learnable scalar ($g$) controlling the correction. 
\begin{equation}
\hat{\mathbf{y}} = \mathbf{y} + \tanh(g)\big[(\tilde{\mathbf{y}}_{\mathrm{lo}} + \tilde{\mathbf{y}}_{\mathrm{hi}}) - \mathbf{y}\big]
\end{equation} 
This operation can be interpreted as an adaptive filter that preserves temporal variation while suppressing broadband noise before the integration operation.  
This component adds $\sim$500 parameters (less than 0.1\% of total parameters) into the model yet improves the performance.
Finally, the BVP waveform $\mathbf{p}_{\text{BVP}}$ is reconstructed by integrating the temporal differences $\hat{\mathbf{y}}$ over time as in Equation~\ref{eq:int}.
\begin{equation}\label{eq:int}
\mathbf{p}_{\text{BVP}} = \sum_{i=1}^{t} \hat{\mathbf{y}}_i, \quad t = 1, \ldots, T
\end{equation}
This integration step recovers the cumulative reflectance variation corresponding to blood flow and serves as the input for the cross-domain pretraining stage.

\begin{figure*}[t]
    \centering
    \includegraphics[width=\linewidth]{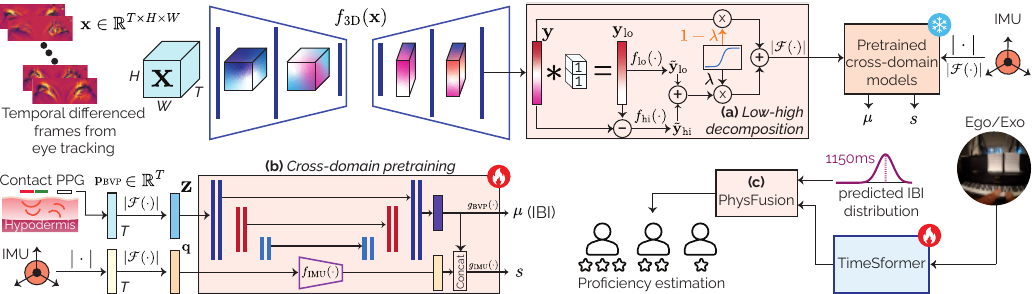}
    \caption{\projname estimates HR and HRV from egocentric gaze videos.  
    Our method introduces three key components: 
    (1)~A low–high decomposition module integrated into the 3D backbone to suppress high-frequency noise and preserve waveform shape.  
    (2)~Cross-domain pretraining of blood volume pulse signals using large-scale physiological datasets  for learning frequency-domain representations to transfer into BVP signals obtained from cameras.  
    (3)~\emph{PhysFusion}, a multimodal fusion head that combines TimeSformer video embeddings with temporally weighted HRV for proficiency estimation in the EgoExo4D benchmark~\cite{ego_exo_4d}.
    }
    \label{fig:placeholder}
\end{figure*}

\subsubsection{Cross-domain pretraining}
\label{sec:crossdomain}
The visual backbone provides a preliminary estimate of the blood volume pulse signal $\mathbf{p}$, which still contains residual noise.  
To improve temporal precision, we leverage large-scale physiological datasets~\cite{wildppg, DeepPPG} containing synchronized contact BVP and electrocardiogram (ECG) recordings.  
From each ECG trace, we extract reference IBIs using Pan–Tompkins~\cite{Pan_Tompkins} by detecting successive cardiac cycles.
Using ECG-derived IBIs as supervision, we train a model to predict IBIs directly from camera BVP segments.

A key challenge in this cross-domain setting is the phase mismatch between contact and vision-based BVP signals caused by differences in sensor location, optical path, and propagation delay~\cite{PPG_regulation, PLOS_PPG_SQ}.  
To mitigate this, we transform each BVP segment into the frequency domain using the discrete Fourier transform (DFT) $\mathcal{F}(\cdot)$ as in Equation~\ref{eq:FFT}.
\begin{equation}\label{eq:FFT}
\mathbf{F} = \mathcal{F}(\mathbf{p}_{\text{BVP}}), \qquad 
\mathbf{z} = |\mathbf{F}| \in \mathbb{R}^{K},
\end{equation}
and use the magnitude spectrum, $\mathbf{z}$, as input to the encoder.  
This representation preserves the frequency structure while discarding phase that varies across domains, which has been shown to be effective for cross-domain BVP signals~\cite{beliefppg}.

The cross-domain BVP encoder, $f_{\text{BVP}}(\cdot)$, uses a three-block U-Net architecture~\cite{Unet}, which has been shown to be effective for this task~\cite{periodicity_detection}, to map $\mathbf{z}$ to a probabilistic IBI distribution. 
To further improve robustness under motion, we integrate inertial features through multimodal fusion.  
For each IMU segment, we first compute the magnitude of the tri-axial acceleration signal and transform it into the frequency domain using the DFT.

The resulting Fourier magnitude spectrum, which captures the dominant motion frequencies and their magnitude, is then processed by a lightweight two-layer convolutional encoder $f_{\text{IMU}}(\mathbf{q})$.  
After decoding the BVP spectrum, we apply global average pooling, which is concatenated with the IMU features and passed through a small MLP head predicting the log-variance $s$ of the distribution.
\begin{equation}\label{eq:prob_ibi}
\mu =  g_{\text{BVP}}(f_{\text{BVP}}(\mathbf{z})) \quad s = g_{\text{IMU}}(f_{\text{BVP}}(\mathbf{z}), f_{\text{IMU}}(\mathbf{q}))
\end{equation}
Detailed architectural configurations (kernel sizes, activation layers, and projection blocks) for cross-domain models are provided in the Appendix.  
The model is trained with a negative log-likelihood (NLL) loss as in Equation~\ref{eq:NLL}.
\begin{equation}\label{eq:NLL}
\textstyle\mathcal{L}_{\text{NLL}} = \frac{1}{N}\sum_i 
\left[\frac{({\mathrm{IBI}}_i - \mu_i)^2}{2e^{s_i}} + \frac{s_i}{2}\right],
\end{equation}
where $(\mu_i, s_i)$ denote the predicted mean and log-variance for each window.  
A training window spans 128~frames, aligned with reference ECG segments, supervised at the window level using the mean IBI.
The mean~$\mu$ provides the estimated mean IBI value, while the variance~$e^{s}$ captures predictive uncertainty.  
This multimodal fusion allows the model to adjust its uncertainty based on motion intensity, reflecting lower confidence during motion-heavy segments.  

At inference, the pretrained encoder is transferred to the visual domain, where it operates on egocentric camera-derived BVP signals to obtain IBI estimates.  


\subsection{PhysFusion for  Proficiency Estimation}
\label{sec:training_prof_estimation}
We introduce \emph{PhysFusion}, a multimodal fusion module that jointly encodes weighted HRVs and video representations for the downstream task of proficiency prediction in egocentric applications.

Given a clip $\mathbf{x}$ from eye tracking cameras, we predict user proficiency by fusing video representations with physiological features.
The video is fed to a TimeSformer~\cite{TimesFormer} to output a class token $\mathbf{v}\in\mathbb{R}^{d}$ for proficiency levels (4-class: novice, early expert, intermediate expert, late expert).
From the pretrained cross-domain model, we obtain per-window IBI features $(\mu_t, s_t)$.
We convert $\mu_t$ [ms] to HR $h_t{=}60{,}000/\mu_t$ and form the physiological feature vector $\mathbf{c}_t{=}[h_t,\, s_t\,]^\top\in\mathbb{R}^{2}$ for each 128~frames.

We build the physiological feature vector for each 128~frames (4.3\,s) in the egocentric eye tracking video $\mathbf{x}$.
Each step is projected with a linear layer and temporally encoded with a transformer as in Equation~\ref{eq:proj_enc}.
\begin{equation}
\label{eq:proj_enc}
\{\mathbf{T}_t\}_{t=1}^{T} = \mathcal{T}_\theta\big(\{\mathbf{u}_t\}_{t=1}^{T}\big), \hspace{2mm} \mathbf{u}_t = \mathbf{W}_p \mathbf{c}_t + \mathbf{b}_p \in \mathbb{R}^{d_p}, 
\end{equation}
where $\mathbf{W}_p,\mathbf{b}_p$ are learnable linear layer, and $\mathcal{T}_\theta$ is a 2-layer, 4-head Transformer encoder (multi-head self-attention with pre-norm residual blocks).
We employ a lightweight encoder to capture temporal dependencies in HRV that are lost under the simple averaging used in prior work~\cite{egoPPG}.
We aggregate the sequence with learned attention as in Equation~\ref{eq:learned_attention}.
\begin{equation}\label{eq:learned_attention}
\textstyle \mathbf{p} = \sum_{t=1}^{T} \alpha_t\, \mathbf{T}_t \in \mathbb{R}^{d_p} \hspace{2mm} \alpha_t = \frac{\exp(\mathbf{w}^\top \mathbf{T}_t)}{\sum_k \exp(\mathbf{w}^\top \mathbf{T}_k)} 
\end{equation}
We designed the projection for heteroscedastic weighting to reduce the contribution of HRVs with larger variance.

Finally, we concatenate the TimeSformer representations with the processed physiological features and apply a linear classification head as in Equation~\ref{eq:concat}.
\begin{equation}\label{eq:concat}
\mathbf{z} = [\mathbf{v};\, \mathbf{p}] \in \mathbb{R}^{d+d_p}, \hspace{3mm} \hat{y} = \mathbf{W}_h \mathbf{z} + \mathbf{b}_h.
\end{equation}
The TimeSformer and PhysFusion modules are trained jointly, while the cross-domain IBI estimator remains frozen.
Gradients flow through both the TimeSformer and PhysFusion modules, while the cross-domain IBI estimator (frozen) produces $(\mu_t, s_t)$ during inference.
Loss is chosen to match the target (classification) and applied to $\hat{y}$.
All projection layers use GELU activations. 
We provide architectural details in Appendix~A.


\section{Experimental setup}
\label{sec:experiments}

We evaluate our method on egoPPG-DB~\cite{egoPPG}, which provides eye-gaze videos, contact-based PPG, and electrocardiograms (ECG) for ground-truth beat timing.

\subsection{Datasets}

\noindent
\textbf{HR \& HRV.}
We used egoPPG-DB~\cite{egoPPG} as a dataset with ground-truth values. 
egoPPG-DB consists of total 13\,hours of recordings from 25~participants performing activities such as kitchen tasks, dancing, and cycling.
Each participant wore Aria glasses equipped with inward-facing eye-tracking cameras (30\,fps, 320$\times$240\,px) and an onboard IMU.
Ground-truth signals were collected using a custom contact PPG sensor located at the nasal bridge and validated against an ECG chest strap.
The diverse activities cause widely ranging motion intensities and HR values (44--164\,bpm).
This diversity provides a realistic test bed for evaluating HR and HRV estimation under naturalistic egocentric conditions.

\vspace{3mm}
\noindent
\textbf{Proficiency estimation.}
To assess the behavioral relevance of the obtained physiological cues, we evaluate our estimated HRV features on the \emph{proficiency estimation} benchmark from the EgoExo4D~\cite{ego_exo_4d}.
EgoExo4D contains over 5{,}000 videos from 740 participants performing skilled human activities such as dancing and sports.
The task aims to classify proficiency (novice, early expert, intermediate expert, late expert) based on egocentric and exocentric video.

Following the previous protocols~\cite{egoPPG}, we integrate the IBIs estimated by our method into the existing video classification pipeline.
Specifically, we obtain IBIs for video clips and concatenate them with the latent features from the TimeSformer~\cite{TimesFormer} before the final classification head.

\subsection{Training setup}

\textbf{HR \& HRV.}
We use the official dataset splits introduced for egoPPG~\cite{egoPPG} and follow a five-fold cross-validation.
Each model was trained on four folds and evaluated on the remaining fold.
Video frames were preprocessed using spatial normalization and temporal differencing to enhance pulsatile signals.
We trained the 3D backbone and low-high decomposition module using mean squared error (MSE) loss between predicted and differentiated reference signals.
Optimization employed the Adam~\cite{Adam} (learning rate $9\times10^{-4}$, batch size 4) for 100 epochs.
All experiments were performed on an NVIDIA RTX~4090 GPU.

\vspace{3mm}
\noindent
\textbf{Proficiency estimation.}
We implement the TimeSformer with the same configuration as for the EgoExo4D~\cite{ego_exo_4d} with a clip size of 16 frames and a sampling rate of 16, trained for 15 epochs.
We use all videos of the EgoExo4D, for which the proficiency labels are available (using the official benchmark training/validation sets) and which have at least 16 frames at a sampling rate of 16, resulting in 2044 videos.
From the official training set, we use 10\% as the held-out validation set for testing, following the exact egoPPG implementation, including upsampling the eyetracking video frame rate to 30 fps through linear interpolation between frames~\cite{egoPPG}.
We use our predicted IBIs and their log-variance for the corresponding videos.
We first train the HR/HRV estimator on egoPPG-DB and then freeze it to extract IBI features ($\mu_t$) and log-variance ($s_t$) for EgoExo4D clips.
We integrate these features using the \emph{PhysFusion} with the TimeSformer’s backbone before feeding it into the classification head. 

\subsection{Evaluation Metrics}
We report results for both HR and HRV.
For HR estimation, we compute the mean absolute error (MAE), root mean squared error (RMSE), mean absolute percentage error (MAPE), and Pearson correlation~($r$) between predicted and reference HR sequences for each 128-frame window.

For HRV evaluation, we used the IBIs estimated by our model and those extracted from the ground-truth ECG signals using the Pan–Tompkins algorithm~\cite{Pan_Tompkins}, and computed standard metrics consistent with HR evaluation.  
Additionally, we report common time-domain HRV indices~\cite{rajendra_acharya_heart_2006}, such as the standard deviation of IBIs (SDNN) and the root mean square of successive differences (RMSSD).  
All HRV measures are compared against ECG-derived reference values.  

For proficiency estimation, we report top-1 classification accuracy following the official EgoExo4D protocol~\cite{ego_exo_4d}.

\subsubsection{Why HRV and not just HR?}

We report HR and HRV every 128 frames (4.3\,s), in addition to the 60\,s evaluation used in prior work.
Although HRV is conventionally computed from beat-to-beat intervals, window-level evaluation is also used in recent biomedical HRV estimation studies in real-world situations~\cite{demirelHRV}.

\vspace{3mm}
\noindent
\textbf{How much information does a 60\,s mean lose?}
Let $H$ denote 4-second HR in bpm.  
Across egoPPG-DB~\cite{egoPPG}, the mean HR is $\mathbb{E}[H]=82.8$\,bpm with $\mathrm{Std}(H)=12.0$\,bpm ($\mathrm{Var}(H)=144$).  
Within 60\,s windows, the average standard deviation is $\bar{\sigma}_{60}=7.6$\,bpm, with $\mathrm{min}=1.1$ and $\mathrm{max}=53.2$\,bpm.  
These statistics are computed from the reference ECG recorded alongside the egocentric videos from EgoPPG~\cite{egoPPG} for accurate ground-truth timing.  
This indicates that intra-minute fluctuations are highly variable across contexts.
Such variability is expected, as the dataset includes dynamic activities such as dancing, biking, and walking, which induce rapid HR changes even within a minute.
This is consistent with our prior findings that cardiac dynamics carry nonlinear temporal structure that minute-level aggregates obscure~\cite{demirel2025temporalcardiovasculardynamics}.

\noindent\\[.1em]
\textbf{How much variance is discarded?}
To quantify the variance lost through 60\,s averaging, as done in prior work~\cite{egoPPG}, we compute how much information is discarded using the law of total variance (Equations~\ref{eq:tot_variance_final_hr} and~\ref{eq:total_variance_2}), highlighting the limitation of such coarse temporal aggregation.
\begin{equation}\label{eq:tot_variance_final_hr}
\mathrm{Var}(H)\!=\!\mathbb{E}[\mathrm{Var}(H\mid\text{minute})]\!+\!\mathrm{Var}(\mathbb{E}[H\mid\text{minute})]
\end{equation}
Thus, minute-level means retain only $\approx 59.8\%$ of the total variance, discarding about $40\%$ of the information.
\begin{equation}\label{eq:total_variance_2}
\frac{\mathrm{Var}(\mathbb{E}[H\mid\text{minute}])}{\mathrm{Var}(H)} = \frac{144 - 57.9}{144} \approx 59.8\%
\end{equation}

\noindent\\[.1em]
\textbf{Implications for egocentric systems.}
Egocentric video captures rapid motions, illumination changes, and viewpoint shifts~\cite{ego_exo_4d} that can modulate HR in little time (5--15\,s).  
Prior egocentric work averages HR over 60\,s windows, which smooths out these fluctuations and discards up to 7--50\,bpm of local error.
\projname instead evaluates HR and HRV continuously at the timescale of these variations and preserves the rapid dynamics that characterize egocentric settings. 

This temporal resolution matters in practice.
rPPG methods assume a stationary camera and subject and therefore lose the within-minute dynamics that egocentric applications must capture for users who move freely during daily life.


\section{Results}
\label{sec:results}

\subsection{HR \& HRV estimation}
Table~\ref{tab:hr_hrv_results} summarizes the results for HR and HRV estimation.  
For HR, we follow prior work and report errors in beats per minute.  
For HRV, we first evaluate the estimated IBIs in milliseconds before reporting the derived time-domain indices.
To provide a comprehensive evaluation, we report results for both short-term ($\sim$4\,s) and long-term (60\,s) windows.
\begin{table*}[htbp]
\centering
\caption{Comparison of heart rate and heart rate variability estimation on egoPPG-DB. 
Metrics are reported at two different time resolutions for HR and IBI: 4\,s and 60\,s. 
SDNN and RMSSD are computed from estimated IBIs over the evaluation windows. 
}
\label{tab:hr_hrv_results}
\vspace{-2mm}
\begin{adjustbox}{width=1\columnwidth,center}
\renewcommand{\arraystretch}{1.1}
\begin{tabular}{lcccccccccccccccccccc}
\toprule
\multirow{3}{*}{\textbf{Method}} 
& \multicolumn{8}{c}{\textbf{HR}} 
& \multicolumn{8}{c}{\textbf{HRV (IBI) [ms]}} 
& \multicolumn{4}{c}{\textbf{HRV Time-Domain [ms]}} \\
\cmidrule(lr){2-9}\cmidrule(lr){10-17}\cmidrule(lr){18-21}
& \multicolumn{2}{c}{\textbf{MAE}} & \multicolumn{2}{c}{\textbf{RMSE}} & \multicolumn{2}{c}{\textbf{MAPE (\%)}} & \multicolumn{2}{c}{\textbf{$r$}}
& \multicolumn{2}{c}{\textbf{MAE}} & \multicolumn{2}{c}{\textbf{RMSE}} & \multicolumn{2}{c}{\textbf{MAPE (\%)}} & \multicolumn{2}{c}{\textbf{$r$}}
& \multicolumn{2}{c}{\textbf{SDNN}} & \multicolumn{2}{c}{\textbf{RMSSD}} \\
\cmidrule(lr){2-3}\cmidrule(lr){4-5}\cmidrule(lr){6-7}\cmidrule(lr){8-9}
\cmidrule(lr){10-11}\cmidrule(lr){12-13}\cmidrule(lr){14-15}\cmidrule(lr){16-17}
\cmidrule(lr){18-19}\cmidrule(lr){20-21}
& \textbf{4\,s} & \textbf{60\,s} & \textbf{4\,s} & \textbf{60\,s} & \textbf{4\,s} & \textbf{60\,s} & \textbf{4\,s} & \textbf{60\,s}
& \textbf{4\,s} & \textbf{60\,s} & \textbf{4\,s} & \textbf{60\,s} & \textbf{4\,s} & \textbf{60\,s} & \textbf{4\,s} & \textbf{60\,s}
& \textbf{MAE} & \textbf{MAPE} & \textbf{MAE} & \textbf{MAPE} \\
\midrule
Baseline eyes & 27.30 & 14.60 & 33.68 & 18.88 & 33.71 & 18.37 & 0.12 & 0.20 & 218.84 & 237.61 & 280.12 & 263.90 & 44.37 & 28.37 & 0.16 & 0.22 & 84.92 & 100.39 & 84.70 & 153.42 \\
Baseline skin & 26.12 & 12.40 & 32.60 & 15.54 & 30.16 & 15.29 & 0.20 & 0.50 & 192.68 & 223.64 & 263.23 & 266.69 & 42.02 & 25.98 & 0.18 & 0.23 & 79.71 & 96.74 & 81.86 & 152.35 \\
PhysNet~\cite{yu2019remote}        & 20.96 & 11.28 & 29.88 & 14.09 & 26.81 & 14.89 & 0.36 & 0.67 & 181.97 & 167.72 & 235.75 & 205.88 & 23.10 & 21.03 & 0.38 & 0.68 & 28.26 & 72.71 & 45.73 & 131.24 \\
PhysFormer~\cite{physformer}     & 16.70 & 13.09 & 21.45 & 15.41 & 19.68 & 16.60 & 0.32 & 0.63 & 130.44 & 105.58 & 178.02 & 145.48 & 17.71 & 15.03 & 0.34 & 0.67 & 29.54 & 73.14 & 47.61 & 129.56 \\
FactorizePhys~\cite{FactorizePhys}  & 14.51 & 13.27 & 19.80 & 16.50 & 16.85 & 17.48 & 0.27 & 0.69 & 200.72 & 210.27 & 275.51 & 250.13 & 27.77 & 26.27 & 0.31 & 0.62 & 28.65 & 72.85 & 45.84 & 129.35 \\
PulseFormer~\cite{egoPPG}    & 22.10 & 8.10 & 29.25 & 11.08 & 27.96 & 10.10 & 0.38 & 0.83 & 147.55 & 123.86 & 203.00 & 160.86 & 19.20 & 15.67 & 0.43 & 0.75 & 33.71 & 85.49 & 53.90 & 146.45  \\
\textbf{Ours}          & \textbf{12.72} & \textbf{7.13} & \textbf{18.27} & \textbf{10.76} & \textbf{14.47} & \textbf{8.73} & \textbf{0.42} & \textbf{0.84}
                       & \textbf{105.91} & \textbf{87.75} & \textbf{149.78} & \textbf{119.80} & \textbf{13.56} & \textbf{12.81} & \textbf{0.54} & \textbf{0.77}
                       & \textbf{10.89} & \textbf{27.05} & \textbf{19.14} & \textbf{42.55} \\
\addlinespace
Gain (\%)  & \textcolor{Green}{12.34} & \textcolor{Green}{11.98} & \textcolor{Green}{7.73} & \textcolor{Green}{2.89} & \textcolor{Green}{14.12} & \textcolor{Green}{13.56} & \textcolor{Green}{10.53} & \textcolor{Green}{1.20} & \textcolor{Green}{18.81} & \textcolor{Green}{16.89} & \textcolor{Green}{15.86} & \textcolor{Green}{17.65} & \textcolor{Green}{23.43} & \textcolor{Green}{14.77} & \textcolor{Green}{25.58} & \textcolor{Green}{2.67}
& \textcolor{Green}{61.46}
& \textcolor{Green}{62.80}
& \textcolor{Green}{58.15}
& \textcolor{Green}{67.10} \\
\bottomrule
\end{tabular}
\end{adjustbox}
\end{table*}

\subsection{HRV time-domain metrics}
Table~\ref{tab:hr_hrv_results} additionally lists SDNN and RMSSD, the most widely adopted measures in practical HRV analysis~\cite{HRV_overview} that quantify overall and short-term variability in beat intervals, respectively.  
SDNN reflects overall variability in IBIs, providing a measure of total autonomic activity, while RMSSD emphasizes short-term fluctuations linked to parasympathetic modulation~\cite{HRV_cardiac_autonomic_system}.  
We focus on time-domain metrics because frequency-domain HRV measures (e.g., LF/HF power ratios) require much longer continuous recordings (typically several hours up to 24\,h~\cite{de_maria_day_2023, HRV_cardiac_autonomic_system}), which current egocentric datasets do not capture.  
We report frequency-domain HRV results for completeness in Appendix~E.1 (Table 9), although these metrics are less reliable for short, activity-rich egocentric clips.

Both time and frequency-domain HRV correlate to stress~\cite{hrv_social_stress}, workload~\cite{HRV_mentalWorkload}, and fatigue~\cite{hrv_fatigue}, making them especially relevant for modeling behavioral state in egocentric settings.  

Our method improves the estimation of these indices by up to \emph{67\%} compared to baselines while these improvements increase when we investigate activities as in Appendix~E.4. 
This shows our method's capability of recovering physiologically meaningful dynamics from short, motion-rich recordings.

\subsection{Proficiency estimation on EgoExo4D}
Table~\ref{tab:proficiency_results} presents the contribution of our HRV estimation to the proficiency prediction benchmark on EgoExo4D.  
Integrating the predicted IBIs and their log variances into TimeSformer improves classification accuracy. 
Our method achieves the highest accuracy in all the activity categories and delivers the best overall performance.  
When combining egocentric videos with our HRV features, the overall accuracy reaches 46.75\%, representing a 17.8\% relative improvement over egocentric video alone.  
These results demonstrate that physiological cues offer complementary behavioral information beyond visual appearance.

\begin{table*}[t]
\centering
\caption{Proficiency estimation results on the EgoExo4D with three random seeds.  
Accuracy (\%) is reported for each scenario and overall for all clips.  
Following PulseFormer (PF)~\cite{egoPPG}, we reproduce their coarse-grained HR (60\,s averaged) with secondary statistics such as mean and standard deviation, while our HRV features are computed every 128 frames and fused with TimeSformer representations using our method (+HRV).
}
\label{tab:proficiency_results}
\vspace{-2mm}
\begin{adjustbox}{width=\columnwidth,center}
\begin{threeparttable}
\renewcommand{\arraystretch}{1.1}
\begin{tabular}{lc ccc | ccc | ccc}
\toprule
\multirow{2}{*}{\textbf{Scenario}}
& \multirow{2}{*}{\textbf{Majority}}
& \multicolumn{3}{c}{\textbf{Ego}}
& \multicolumn{3}{c}{\textbf{Exo}}
& \multicolumn{3}{c}{\textbf{Ego + Exo}} \\
\cmidrule(lr){3-5}\cmidrule(lr){6-8}\cmidrule(lr){9-11}
& & Baseline & +HR (PF)  & +HRV (ours)
  & Baseline & +HR (PF)  & +HRV (ours)
  & Baseline & +HR (PF)  & +HRV (ours) \\
\midrule
Basketball & 38.00 
& 45.45\scriptsize$\pm$0.62 & 47.47\scriptsize$\pm$0.58 & \textbf{48.48}\scriptsize$\pm$0.55
& 48.48\scriptsize$\pm$0.60 & \textbf{48.98}\scriptsize$\pm$0.57 & 47.67\scriptsize$\pm$0.62
& 49.49\scriptsize$\pm$0.59 & 50.10\scriptsize$\pm$0.55 & \textbf{52.52}\scriptsize$\pm$0.53 \\
Cooking    &  0.00 
& 20.00\scriptsize$\pm$0.70 & 20.00\scriptsize$\pm$0.68 & \textbf{25.00}\scriptsize$\pm$0.65
& 35.00\scriptsize$\pm$0.63 & 33.75\scriptsize$\pm$0.66 & \textbf{36.00}\scriptsize$\pm$0.62
& 25.00\scriptsize$\pm$0.69 & \textbf{35.50}\scriptsize$\pm$0.61 & 35.00\scriptsize$\pm$0.60 \\
Dancing    & 24.59 
& 43.44\scriptsize$\pm$0.64 & 44.26\scriptsize$\pm$0.60 & \textbf{47.54}\scriptsize$\pm$0.58
& 42.62\scriptsize$\pm$0.66 & 42.62\scriptsize$\pm$0.65 & \textbf{47.54}\scriptsize$\pm$0.60
& 50.82\scriptsize$\pm$0.62 & 55.73\scriptsize$\pm$0.59 & \textbf{60.00}\scriptsize$\pm$0.57 \\
Music      & 57.89 
& 78.94\scriptsize$\pm$0.48 & 77.50\scriptsize$\pm$0.50 & \textbf{79.78}\scriptsize$\pm$0.46
& 57.89\scriptsize$\pm$0.55 & \textbf{62.50}\scriptsize$\pm$0.53 & 52.10\scriptsize$\pm$0.60
& 57.89\scriptsize$\pm$0.54 & 60.13\scriptsize$\pm$0.52 & \textbf{60.52}\scriptsize$\pm$0.50 \\
Bouldering & 15.29 
& 24.50\scriptsize$\pm$0.78 & 30.37\scriptsize$\pm$0.72 & \textbf{39.89}\scriptsize$\pm$0.70
&  8.61\scriptsize$\pm$0.60 & 14.54\scriptsize$\pm$0.58 & \textbf{16.99}\scriptsize$\pm$0.56
& 14.64\scriptsize$\pm$0.74 & 18.75\scriptsize$\pm$0.70 & \textbf{25.49}\scriptsize$\pm$0.66 \\
Soccer\textsuperscript{a}    & 62.50 
& 50.00\scriptsize$\pm$8.62 & 60.00\scriptsize$\pm$6.35 & \textbf{67.50}\scriptsize$\pm$7.78
& \textbf{81.25}\scriptsize$\pm$7.11 & 75.00\scriptsize$\pm$8.26 & 68.75\scriptsize$\pm$7.59
& \textbf{75.75}\scriptsize$\pm$7.44 & 62.50\scriptsize$\pm$8.52 & 70.80\scriptsize$\pm$7.67 \\
\midrule
\textbf{Overall} & 27.80 
& 39.69\scriptsize$\pm$0.41 & 44.30\scriptsize$\pm$0.38 & \textbf{46.75}\scriptsize$\pm$0.36
& 34.75\scriptsize$\pm$0.43 & 37.00\scriptsize$\pm$0.40 & \textbf{37.77}\scriptsize$\pm$0.39
& 37.80\scriptsize$\pm$0.42 & 41.18\scriptsize$\pm$0.39 & \textbf{43.52}\scriptsize$\pm$0.37 \\
\bottomrule
\end{tabular}
\begin{tablenotes}
\footnotesize
\item[a] Underrepresented scenario (70 out of $\sim$2k samples), contributing to higher variance.
\end{tablenotes}
\end{threeparttable}
\end{adjustbox}
\end{table*}

\subsection{Ablation experiments}
\label{sec:ablation}

\noindent
\textbf{HR \& HRV estimation ablation.}
Table~\ref{tab:ablation_hr_hrv} presents the results.  
The baseline model (3D backbone) predicts temporal differences using standard differencing.  
Introducing the low–high decomposition substantially reduces HR error.  
Finally, incorporating cross-domain pretraining yields the largest overall gain, demonstrating the benefit of transferring representations learned from contact BVP-ECG data.  
The full model achieves the best HR and HRV accuracy across all metrics.

\begin{table*}[t]
\centering
\caption{
Ablation study on HR and HRV estimation.
Each component is added cumulatively; ``--'' indicates the exclusion
of a design from cross-domain pretraining.
The backbone is a 3D CNN~\cite{yu2019remote} with spatial
attention~\cite{CBAM}, applied to eye-tracking video~\cite{egoPPG}.
}
\label{tab:ablation_hr_hrv}
\vspace{-2mm}
\begin{adjustbox}{max width=\textwidth,center}
\small
\setlength{\tabcolsep}{3pt}
\renewcommand{\arraystretch}{1.1}

\begin{tabular}{@{}l*{14}{c}@{}}
\toprule
\multirow{3}{*}{\textbf{Configuration}}
& \multicolumn{4}{c}{\textbf{HR (bpm)}}
& \multicolumn{4}{c}{\textbf{HRV (IBI, ms)}}
& \multicolumn{6}{c}{\textbf{HRV Time Indices (ms)}} \\
\cmidrule(lr){2-5}
\cmidrule(lr){6-9}
\cmidrule(lr){10-15}

& \multirow{2}{*}{MAE}
& \multirow{2}{*}{RMSE}
& \multirow{2}{*}{\mapehead}
& \multirow{2}{*}{$r$}
& \multirow{2}{*}{MAE}
& \multirow{2}{*}{RMSE}
& \multirow{2}{*}{\mapehead}
& \multirow{2}{*}{$r$}
& \multicolumn{3}{c}{\textbf{SDNN}}
& \multicolumn{3}{c}{\textbf{RMSSD}} \\
\cmidrule(lr){10-12}
\cmidrule(lr){13-15}

& & & &
& & & &
& MAE & RMSE & \mapehead
& MAE & RMSE & \mapehead \\
\midrule

Baseline (3D Backbone)
& 21.76 & 29.25 & 27.96 & 0.38
& 152.61 & 209.97 & 19.55 & 0.42
& 31.27 & 35.73 & 63.75
& 48.07 & 54.20 & 93.25 \\

+ Low--High decomposition
& 14.80 & 20.46 & 18.10 & 0.40
& 117.72 & 167.99 & 15.93 & 0.44
& 26.96 & 27.66 & 35.35
& 34.72 & 38.94 & 55.61 \\

+ Cross-domain pretraining (\textbf{ours})
& \textbf{12.72} & \textbf{18.27}
& \textbf{14.47} & \textbf{0.42}
& \textbf{105.91} & \textbf{149.78}
& \textbf{13.56} & \textbf{0.54}
& \textbf{10.89} & \textbf{16.12}
& \textbf{27.05}
& \textbf{19.14} & \textbf{26.13}
& \textbf{42.55} \\

\hspace{3mm}-- Fourier magnitude
& 14.87 & 20.51 & 16.72 & 0.31
& 130.30 & 173.57 & 19.23 & 0.33
& 12.38 & 18.91 & 31.28
& 20.48 & 27.83 & 48.37 \\

\hspace{3mm}-- NLL objective
& 15.34 & 20.09 & 18.81 & 0.08
& 133.47 & 179.33 & 18.14 & 0.10
& 28.36 & 33.08 & 47.59
& 29.89 & 36.81 & 46.15 \\

\hspace{3mm}-- U-Net (use 3-layer MLP)
& 13.40 & 19.12 & 15.74 & 0.08
& 111.34 & 160.82 & 16.62 & 0.51
& 11.91 & 17.29 & 28.16
& 20.42 & 26.93 & 44.30 \\

\bottomrule
\end{tabular}
\end{adjustbox}
\end{table*}

\noindent\\[.5em]
\textbf{Proficiency estimation ablation.}
To assess the impact of HRV estimation, we analyze how different HRV representations affect proficiency prediction accuracy on the EgoExo4D benchmark.  
Table~\ref{tab:ablation_proficiency} reports results for three variants. 
The first variant (+ HR) employs 60\,s averaged HR values.
The second uses HRV features without log variances and the third incorporates probabilistic HRV features trained with the NLL objective to model uncertainty with our method (the uncertainty evaluations with a calibration plot are given in Appendix Section~E.3). 

\begin{table}[t]
\centering
\caption{Ablation on proficiency estimation using EgoExo4D.
Top-1 accuracy (\%) per activity and overall.
Columns use compact labels due to space constraints: Ba=Basketball, Co=Cooking, Da=Dancing, Mu=Music, Bo=Bouldering, So=Soccer.}
\label{tab:ablation_proficiency}
\vspace{-2mm}
\renewcommand{\arraystretch}{1.1}
\begin{adjustbox}{width=0.8\columnwidth,center}
\begin{tabular}{lccccccc}
\toprule
\textbf{Configuration} & \textbf{Ba} & \textbf{Co} & \textbf{Da} & \textbf{Mu} & \textbf{Bo} & \textbf{So} & \textbf{Overall} \\
\midrule
Baseline                                   & 45.45 & 20.00 & 43.44 & 78.94 & 24.50 & 50.00 & 39.69 \\
+ HR (60\,s mean)                  & 47.47 & 20.00 & 44.26 & 77.50 & 30.37 & 60.00 & 44.30 \\
+ HRV (deterministic, 4\,s)                & 46.50 & 25.00 & 46.72 & 75.31 & 37.25 & 31.25 & 45.31 \\
+ HRV (uncertainty-aware, \textbf{ours})  & \textbf{48.48} & \textbf{25.00} & \textbf{47.54} & \textbf{79.78} & \textbf{39.89} & \textbf{67.50} & \textbf{46.75} \\
\bottomrule
\end{tabular}
\end{adjustbox}
\end{table}


Introducing probabilistic modeling improves accuracy.  
Using HRV features yields the largest performance gain, indicating that inter-beat intervals carry more discriminative information about user proficiency than HR trends.  
These results demonstrate that the physiological representations learned through cross-domain pretraining enhance downstream reasoning about human skill and behavior.

\subsection{Uncertainty modeling}
\begin{figure}[t]
    \centering
    \begin{minipage}[t]{0.46\columnwidth}
        \centering
        \includegraphics[width=0.9\linewidth]
        {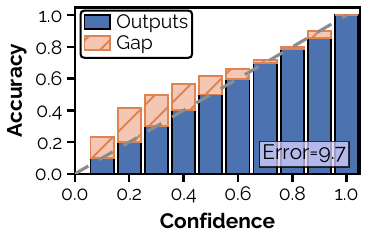}
        \\[-1mm]
        {\small (a) With IMU}
    \end{minipage}
    \hfill
    \begin{minipage}[t]{0.46\columnwidth}
        \centering
        \includegraphics[width=0.9\linewidth]
        {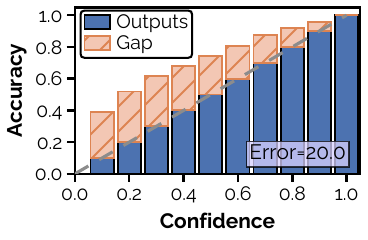}
        \\[-1mm]
        {\small (b) Without IMU}
    \end{minipage}
    \vspace{-2mm}
    \caption{
    Reliability diagrams for uncertainty estimation with and without
    IMU input. 
    }
    \label{fig:imu_reliability}
\end{figure}

We evaluate the contribution of predictive uncertainty and IMU separately.
First, removing the uncertainty features from PhysFusion does not change the HRV point estimates, but reduces proficiency accuracy,
showing that the predicted uncertainty provides additional information for downstream fusion.
Second, we evaluate the role of IMU in uncertainty calibration.
Removing the IMU increases the expected calibration error (ECE) from 9.7\% to 20.0\% (Fig.~\ref{fig:imu_reliability}). 
The IMU provides a motion cue that is independent of visual appearance and skin visibility.

\subsection{Efficiency}
\label{sec:efficiency}

Our HR/HRV pipeline runs online on a single GPU and stays faster than real time on CPU.
The model has 1\,M parameters and requires 40.80\,GFLOPs per 128-frame (4.3\,s) window.
On an NVIDIA RTX~4090, inference takes 5.75\,ms per window with 74\,MB peak memory, a real-time factor of $748\times$.
A 10-thread CPU still reaches $9.6\times$ real time.
Full timings are reported in the appendix (Table~8).


\section{Discussion}
\label{sec:discussion}
\textbf{Performance \& implications.}
Our results show that our method accurately estimates HR and HRV from gaze videos on egocentric vision systems.
Our method outperforms prior HR estimation approaches. 
Our results set the state of the art across both HR and HRV, reducing error in HRV indices, SDNN and RMSSD, by about 65\% relative to the best baselines on egoPPG-DB.

Despite this step forward in accuracy, there is room for improvement in HRV indices that depend on the standard deviation of IBIs.  
Because these measures are sensitive to even a \emph{single} wrongly estimated IBI, local errors can disproportionately affect overall variability.  
Future work could leverage the predicted log-variance to adaptively downweight uncertain intervals to stabilize HRV metric computation, similar to our proposed \emph{PhysFusion} mechanism.

\vspace{3mm}
\noindent
\textbf{Relevance of uncertainty-aware HRV for human behavior modeling.}
Integrating HRV into the EgoExo4D proficiency benchmark resulted in accuracy gains, with the largest improvements observed in physically or cognitively demanding tasks such as bouldering.
Our ablation study shows that our integration of uncertainty estimation accounts for the largest proportional improvement.
Further analysis in Appendix~E.3 shows that the predicted uncertainty is calibrated, correlates with estimation error, and enables unreliable windows to be down-weighted or excluded.
These results suggest that IBIs provide complementary information about effort, attention, and skill assessment.  


\vspace{3mm}
\noindent
\textbf{Towards stress- and arousal-aware egocentric systems.}
HRV is a well-established noninvasive proxy for autonomic stress responses~\cite{thayer2012metaanalysis, kim2018stress, castaldo2015acute}.
The same continuous HRV signal that drives our proficiency gains can thus feed downstream stress- and arousal-aware applications, such as fatigue and workload monitoring in safety-critical activities or affect-sensitive interaction in Mixed Reality.

\vspace{3mm}
\noindent
\textbf{Limitations.}
First, the dataset we built on (egoPPG-DB) lacks outdoor recordings, such that our evaluation does not cover weather variability or broader activities.
Second, we treat HRV as a stress-related autonomic marker rather than a direct stress label, in line with standard HRV interpretation guidance~\cite{taskforce1996hrv, shaffer2017overview, laborde2017hrv}.
We leave a dedicated stress-benchmark evaluation to future work.
Finally, it remains to be tested how well our learned HRV representations transfer to additional downstream tasks (e.g., workload estimation).


\section{Conclusion}
\label{sec:conclusion}
We have presented a method for estimating heart rate and heart rate variability from gaze videos on egocentric vision systems.
HRV extends the capacity of egocentric vision models for behavioral understanding much beyond the heart rate tracking in previous work.
Using public datasets, our model accurately estimates physiological metrics even under substantial head motion and dynamic illumination changes.
We have also demonstrated that our HRV estimates improve downstream user proficiency prediction on the EgoExo4D benchmark by a relative 17.8\%, which underscores the behavioral relevance of heart rate variability for modeling skill and better understanding human behavior more generally.
Since HRV is an established autonomic proxy for perceived stress, our estimates therefore enables stress- and arousal-aware first-person applications without additional body-worn sensors.
Our results position HRV as a core physiological signal for egocentric systems, linking observable behavior with internal physiological state and advancing the foundation for behavioral AI.
Our uncertainty-aware method also enhances reliability by quantifying confidence in physiological estimates, which affords egocentric systems more trustworthy behavioral inference in real-world settings.
Integrating richer physiological representations into multimodal models in future work could enable deeper behavioral inference to jointly capture intent, attention, and adaptation in real-world settings.

{
\makeatletter
\let\oldurl\url
\renewcommand{\url}[1]{}
\let\olddoi\doi
\renewcommand{\doi}[1]{}
\bibliographystyle{splncs04}
\bibliography{main}
\let\url\oldurl
\let\doi\olddoi
\makeatother
}


\clearpage
\setcounter{page}{1}
\appendix

\section{Architectures}
\label{sec:appendix_arch}
In this section, we give details about the architectures that are used for our method.
Since our 3D backbone is adapted from previous works, we have reported the low-high decomposition, cross-domain pretraining models and PhysFusion.

\subsection{Low-high decomposition}

We integrate a lightweight low–high frequency decomposition block into the 3D backbone to suppress broadband motion noise while preserving pulsatile dynamics.
This module operates directly on the temporal feature sequence $\mathbf{y}\in\mathbb{R}^{B\times1\times T}$ and introduces fewer than 0.1\% additional parameters to the overall model.

\begin{table}[h]
\centering
\footnotesize
\setlength{\tabcolsep}{6pt}
\renewcommand{\arraystretch}{1.15}
\caption{Low-high decomposition block details.}
\begin{tabularx}{\columnwidth}{@{}lX@{}}
\toprule
\textbf{Component} & \textbf{Specification} \\
\midrule
Input & $\mathbf y\in\mathbb{R}^{B\times1\times T}$ (from 3D backbone). \\

Low-pass & Box filter (AvgPool/Conv) with $k{=}11$, stride $1$, pad $5$:
$\ \mathbf y_{\mathrm{lo}}=\mathbf y*h,\ h=\tfrac{1}{11}\mathbf 1_{11}$. \\

High-pass & $\mathbf y_{\mathrm{hi}}=\mathbf y-\mathbf y_{\mathrm{lo}}$. \\

Head$_{\mathrm{lo}}$ & Conv1D $1{\rightarrow}16$ ($k{=}9$, $p{=}4$) $\rightarrow$ GELU $\rightarrow$ Conv1D $16{\rightarrow}1$ ($k{=}1$).  
Output $\tilde{\mathbf y}_{\mathrm{lo}}\in\mathbb{R}^{B\times1\times T}$. \\

Head$_{\mathrm{hi}}$ & Conv1D $1{\rightarrow}16$ ($k{=}5$, $p{=}2$) $\rightarrow$ GELU $\rightarrow$ Conv1D $16{\rightarrow}1$ ($k{=}1$).  
Output $\tilde{\mathbf y}_{\mathrm{hi}}\in\mathbb{R}^{B\times1\times T}$. \\

Fusion gain & Learnable scalar $g$ (initialized to $0.3$). \\

Residual fusion & $\hat{\mathbf y}=\mathbf y+\tanh(g)\big[(\tilde{\mathbf y}_{\mathrm{lo}}+\tilde{\mathbf y}_{\mathrm{hi}})-\mathbf y\big]$ (identity as $g{\to}0$). \\

Output & $\hat{\mathbf y}\in\mathbb{R}^{B\times1\times T}$. \\

Params & $\sim\!300$ total (two heads + gate). \\
\bottomrule
\end{tabularx}

\vspace{3pt}
\raggedright\footnotesize
\textit{Notes.} $k$: kernel size, $p$: padding.  
All convolutions use stride $1$ with "same" padding.  
The low-pass kernel ($k{=}11$) spans $\approx0.37$\,s at 30\,Hz.  
\end{table}

\subsection{Cross-domain models}
To enhance the performance of egocentric IR cameras in extracting physiological indicators, we leverage large-scale datasets from contact-based sensors that record blood volume pulse (BVP) and inertial data. 
The architecture used for this cross-domain pretraining is detailed in Table~\ref{tab:cross_domain}.

The model first processes the Fourier magnitude of the integrated BVP signals using a U-Net encoder–decoder to extract temporal features. 
In parallel, the Fourier magnitude of the inertial signal magnitude is processed by a lightweight convolutional encoder to capture motion dynamics. 
The BVP encoder predicts the mean IBI, while the fused BVP and IMU representations are used to predict the log-variance.

The overall network contains approximately 230k parameters in the multimodal configuration.  
During proficiency prediction, the cross-domain model is kept frozen and used only as a feature extractor.

\begin{table}[h]
\centering
\footnotesize
\setlength{\tabcolsep}{6pt}
\renewcommand{\arraystretch}{1.15}
\caption{\label{tab:cross_domain}Architectural details of the cross-domain models used for estimating IBIs. 
All inputs are Fourier magnitude features. $k$: kernel size, $p$: padding.}
\begin{tabularx}{\columnwidth}{@{}lX@{}}
\toprule
\textbf{Component} & \textbf{Specification} \\
\midrule
\multicolumn{2}{l}{\textbf{BVP encoder--decoder ($f_{\text{BVP}}$)}} \\[1pt]
Input & $\mathbf{z}\in\mathbb{R}^{B\times1\times128}$ (FFT magnitude of BVP segment). \\

Downsampling & Four sequential blocks:
\newline (1) Conv1D $1{\rightarrow}N$ ($k{=}3$, $p{=}1$) $\rightarrow$ BN $\rightarrow$ ReLU; 
\newline (2) Conv1D $N{\rightarrow}2N$ ($k{=}3$, $p{=}1$) $\rightarrow$ BN $\rightarrow$ ReLU; 
\newline (3--4) additional down paths with concatenated pooled inputs (kernel $3$, stride $2$). \\

Upsampling & Three symmetric decoder stages with linear interpolation (scale factor 2) followed by concatenation and Conv1D–BN–ReLU blocks ($k{=}3$, $p{=}1$). \\

Output (mean head) & Conv1D $N{\rightarrow}1$ ($k{=}3$, $p{=}1$) $\rightarrow$ Linear (128$\rightarrow$1) $\Rightarrow \mu$. \\

Global pooling & AdaptiveAvgPool1D(1) $\rightarrow$ Linear($N{\rightarrow}128$) to form BVP embedding. \\

\midrule
\multicolumn{2}{l}{\textbf{IMU encoder ($f_{\text{IMU}}$)}} \\[1pt]
Input & $\mathbf{q}\in\mathbb{R}^{B\times1\times128}$ (FFT magnitude of IMU magnitude signal). \\

Conv blocks & Conv1D $1{\rightarrow}64$ ($k{=}9$, $p{=}4$) $\rightarrow$ GELU $\rightarrow$ Conv1D $64{\rightarrow}64$ ($k{=}9$, $p{=}4$) $\rightarrow$ GELU $\rightarrow$ AdaptiveAvgPool1D(1). \\

Projection & Linear($64{\rightarrow}64$) $\Rightarrow$ IMU feature. \\

\midrule
\multicolumn{2}{l}{\textbf{Fusion and probabilistic output}} \\[1pt]
Fusion & Concatenate BVP (128-D) and IMU (64-D) embeddings $\rightarrow$ 192-D vector. \\

Log-variance head (IMU) & MLP: Linear(192$\rightarrow$64) $\rightarrow$ GELU $\rightarrow$ Linear(64$\rightarrow$1) $\Rightarrow s$. \\

Output & Predicted IBI mean $\mu$ and log-variance $s$; trained with NLL loss. \\

Params & $\sim227$\,k (with IMU fusion), $\sim210$\,k (PPG-only). \\

\bottomrule
\end{tabularx}

\vspace{3pt}
\raggedright\footnotesize
\textit{Notes.} 
$N{=}32$ channels in the base layer. 
All convolutions use stride $1$ and ``same'' padding.
FFT input length $K{=}128$ corresponds to 4.3\,s temporal windows at 30\,Hz.
\end{table}

\subsection{PhysFusion}
We fuse video features with per-timestep physiological cues.
We mainly use the IBI estimations while normalizing them into per minute values and model log-variance ($\log\sigma_t^2$). 
At each time step we stack $[h_t,\, \log\sigma_t^2]$, project with a linear layer to a 128-D embedding, and encode the sequence with a 2-layer Transformer. 
We apply a learned attention pooling over time to get a single 128-D physiology vector. 
This vector is concatenated with the ViT CLS token and passed to a linear head for prediction.
The details of the model are given in Table~\ref{tab:physfusion}.

\begin{table}[t]
\centering
\footnotesize
\setlength{\tabcolsep}{6pt}
\renewcommand{\arraystretch}{1.12}
\caption{The proposed PhysFusion block details (video CLS fused with physiological stream).}
\label{tab:physfusion}
\begin{tabularx}{\columnwidth}{@{}lX@{}}
\toprule
\textbf{Component} & \textbf{Specification} \\
\midrule
Inputs & Per-timestep HR $h_t$ and log-variance $\log\sigma_t^2$; tensors $h,\log\sigma^2\in\mathbb{R}^{B\times T}$. Stack to $[B,T,2]$. \\

Projection & Linear $(2\rightarrow128)$ applied at each time step $\Rightarrow$ $[B,T,128]$ (\(d_{\text{phys}}=128\)). \\

Temporal encoder & TransformerEncoder, 2 layers; each layer: \(d_{\text{model}}{=}128\), \texttt{nhead}$=4$, \texttt{dim\_feedforward}$=256$, \texttt{dropout}$=0.1$, \texttt{batch\_first}$=$True. \\

Attention pooling & Learned scores via Linear $(128\rightarrow1)$; masked softmax over $T$; weighted sum $\Rightarrow$ physiology vector $v_{\text{phys}}\in\mathbb{R}^{B\times128}$. \\

Fusion & Concatenate ViT CLS ($\in\mathbb{R}^{B\times \text{embed\_dim}}$) with $v_{\text{phys}}$ $\Rightarrow$ $[B,\text{embed\_dim}{+}128]$. \\

Head & Linear $(\text{embed\_dim}{+}128 \rightarrow \text{num\_classes})$ for prediction. \\
\bottomrule
\end{tabularx}

\vspace{3pt}
\raggedright\footnotesize
\textit{Notes.} The cross-domain encoder is kept frozen during the training of PhysFusion, which learns end-to-end with the classifier.
\end{table}

\section{Training details}

\subsection{Low-high decomposition}
The low–high decomposition module is trained jointly with the 3D backbone following the same training setup as in prior work~\cite{egoPPG}. 
Training uses the Adam optimizer with a learning rate of $9{\times}10^{-4}$, batch size of 4, and runs for 100 epochs.
The best model is chosen with the lowest MAE.

\subsection{Cross-domain pretraining}
We pretrain the cross-domain encoder on large-scale contact-sensor datasets using the Fourier magnitude inputs.
Each training window spans $T{=}128$ samples, with supervision provided by ECG-derived IBIs extracted using the Pan–Tompkins algorithm. 

The Dalia dataset is preprocessed with a second-order Butterworth band-pass filter (0.5–8\,Hz) applied to the PPG signal. 
The IMU magnitude is computed and both signals are resampled to match the temporal specifications of the Aria in egoPPG dataset~\cite{egoPPG} and the 3D backbone ($T{=}128$ samples per window). 
After preprocessing, the magnitude spectra from the discrete Fourier transform are used as input to the BVP and IMU encoders.

The training of the cross-domain models is performed using the Adam optimizer~\cite{Adam} with a learning rate of $5{\times}10^{-4}$, weight decay of $1{\times}10^{-7}$, batch size of 128.
We train for 100 epochs.
We also use a learning scheduler adjusts the learning rate based on validation MAE (factor 0.5, patience 15, minimum learning rate $1{\times}10^{-6}$), and early stopping is applied using the same criterion. 
The model checkpoint with the lowest validation MAE is selected and reloaded for usage in egocentric setups without requiring further finetuning.
After pretraining, the cross-domain encoder is frozen and used as a fixed feature extractor for proficiency prediction.


\section{Efficiency report}
\label{sec:appendix_efficiency}

We evaluate the computational cost of the HR/HRV pipeline on a 128-frame (4.3\,s) input window.
The model has approximately 1\,M parameters and requires 20.40\,B multiply--accumulate operations (40.80\,GFLOPs) per window.

Table~\ref{tab:eff} reports inference latency on GPU and CPU.
On an NVIDIA RTX~4090, the model processes one window in 5.75\,ms, or roughly 174 windows per second.
This is about 748 times faster than the 4.3\,s acquisition duration of the input window, and requires only 74\,MB of peak GPU memory.
On CPU, inference takes 447.7\,ms with 10 threads, or 2.23 windows per second and a real-time factor of approximately $9.6\times$.

These results show that our method supports online inference on a GPU and stays faster than real time on a multi-threaded CPU.
The low parameter count and memory footprint also make the model suitable for continuous egocentric sensing pipelines.

\begin{table}[h]
\centering
\vspace{-2mm}
\caption{Inference efficiency for a 128-frame (4.3\,s) input window. The real-time factor is the input duration divided by inference latency.}
\label{tab:eff}
\vspace{-2mm}
\setlength{\tabcolsep}{4pt}
\renewcommand{\arraystretch}{1.05}
\begin{tabular}{@{}lrrrr@{}}
\toprule
Device
& Time (ms)
& Windows/s
& Real-time factor
& Peak mem. (MB) \\
\midrule
RTX~4090
& 5.75
& 173.9
& $747.8\times$
& 74 \\
CPU (10 threads)
& 447.7
& 2.23
& $9.6\times$
& -- \\
\bottomrule
\end{tabular}
\vspace{-2mm}
\end{table}

\section{Reproducibility}
For HR estimates, we followed the official implementation and evaluation settings provided in the \emph{egoPPG} repository
(\url{https://github.com/eth-siplab/egoPPG}).
We note that our reproduced PulseFormer results differ slightly from the values reported in the original \emph{egoPPG} paper. The original paper reports results from a single training run, whereas the repository additionally reports the mean and standard deviation across three random seeds and notes variability across seeds and execution environments.
Our reproduced values are close to the range reported by this multi-seed evaluation.

For proficiency estimation, we used the official TimeSformer implementation
(\url{https://github.com/facebookresearch/TimeSformer})
with its default hyperparameter configuration. All additional modifications related to HRV integration and fusion are described in the main paper and supplementary Sections~\ref{sec:method}--\ref{sec:training_prof_estimation}.

\section{Additional results}

\subsection{Frequency domain based HRV indices}
\label{sec:appendix_freq}
In this section, we have provided additional results regarding frequency-domain based HRV indices derived from estimated IBIs.
Table~\ref{tab:freq_hrv_results} presents the results for LF. HF and LF/HF ratio while comparing our method with previous techniques. 
As can be seen from the table, our method improves the performance up to 31.8\% compared to previous works.

\begin{table}[h]
\centering
\caption{
Frequency-domain HRV indices computed from estimated IBIs.  
LF (0.04–0.15\,Hz) and HF (0.15–0.40\,Hz) are reported in ms$^2$; LF/HF is unitless.  
These metrics are included for completeness, though short egocentric clips make them less reliable.
}
\label{tab:freq_hrv_results}
\renewcommand{\arraystretch}{1.05}
\begin{adjustbox}{width=\columnwidth,center}
\begin{tabular}{lcccccc}
\toprule
\multirow{2}{*}{\textbf{Method}} 
& \multicolumn{2}{c}{\textbf{LF (ms$^2$)}} 
& \multicolumn{2}{c}{\textbf{HF (ms$^2$)}} 
& \multicolumn{2}{c}{\textbf{LF/HF Ratio}} \\
\cmidrule(lr){2-3}\cmidrule(lr){4-5}\cmidrule(lr){6-7}
& MAE $\downarrow$ & $r$ $\uparrow$
& MAE $\downarrow$ & $r$ $\uparrow$
& MAE $\downarrow$ & $r$ $\uparrow$ \\
\midrule
PhysNet~\cite{yu2019remote}        & 310.8 & 0.20 & 278.2 & 0.20 & 0.42 & 0.18 \\
PhysFormer~\cite{physformer}       & 285.5 & 0.22 & 252.4 & 0.22 & 0.38 & 0.20 \\
FactorizePhys~\cite{FactorizePhys} & 298.1 & 0.21 & 269.9 & 0.21 & 0.40 & 0.19 \\
PulseFormer~\cite{egoPPG}          & 271.6 & 0.25 & 246.1 & 0.25 & 0.35 & 0.22 \\
\textbf{Ours}                      & \textbf{209.7} & \textbf{0.30} & \textbf{197.4} & \textbf{0.31} & \textbf{0.29} & \textbf{0.29} \\
\addlinespace
\textbf{Gain (\%)} 
& \textcolor{Green}{\textbf{22.8}} & \textcolor{Green}{\textbf{20.0}} 
& \textcolor{Green}{\textbf{19.8}} & \textcolor{Green}{\textbf{24.0}} 
& \textcolor{Green}{\textbf{17.1}} & \textcolor{Green}{\textbf{31.8}} \\
\bottomrule
\end{tabular}
\end{adjustbox}
\end{table}
Beyond these improvements, frequency-domain HRV metrics still rely on relatively long and stable windows, which limits their robustness in short egocentric clips with rapid motion.  
While our method already raises the accuracy and correlation across LF, HF, and LF/HF, there is room for progress.  
Future work can model temporal structure with sequence-level representations or architectures that capture transient patterns in IBIs.  
Such approaches may recover the fast dynamics, giving more reliable estimates and better downstream performance in egocentric settings.

\subsubsection{60\,s HRV}
Since we reported the performance of 4\,s HRV in Table~1, we also performed additional experiments to show the advantage of higher resolution HRV in the proficiency estimation downstream task.
Table~\ref{tab:proficiency_ego_hrv_60s} presents the results.
\begin{table}[h]
\centering
\caption{Proficiency estimation accuracy (\%) on EgoExo4D (ego-view only). 
We also reported the coarse-grained HRV statistics averaged over 60\,s, whereas our main method uses HRV features computed every 128 frames.}
\label{tab:proficiency_ego_hrv_60s}
\begin{adjustbox}{width=0.8\columnwidth,center}
\renewcommand{\arraystretch}{1.1}
\begin{tabular}{lccccc}
\toprule
\multirow{2}{*}{\textbf{Scenario}} 
& \multirow{2}{*}{\textbf{Majority}} 
& \multirow{2}{*}{\textbf{Baseline}} 
& \multicolumn{2}{c}{\textbf{+HRV}} \\
\cmidrule(lr){4-5}
& & & \textbf{(60\,s, ours)} & \textbf{(4\,s, ours)} \\
\midrule
Basketball & 38.00 & 45.45 & 40.00 & \textbf{48.48} \\
Cooking    &  0.00 & 20.00 & 20.00 & \textbf{25.00} \\
Dancing    & 24.59 & 43.44 & 40.59 & \textbf{47.54} \\
Music      & 57.89 & 78.94 & 77.50 & \textbf{79.78} \\
Bouldering & 15.29 & 24.50 & 20.30 & \textbf{39.89} \\
Soccer     & 62.50 & 50.00 & 60.00 & \textbf{67.50} \\
\midrule
\textbf{Overall} & 27.80 & 39.69 & 43.50 & \textbf{46.75} \\
\bottomrule
\end{tabular}
\end{adjustbox}
\end{table}

The 60\,s variant captures slow trends but loses short-term changes tied to motion and skill execution.  
As in Table~\ref{tab:proficiency_ego_hrv_60s}, the higher-resolution representation gives better accuracy. 
This supports our main claim that preserving HRV dynamics is important for proficiency estimation, where local performance cues often appear within a few seconds rather than across a full minute.

The results align with the variance analysis in the main script.
Coarse aggregation removes a notable part of the signal, while our method keeps within-minute variability that correlates with behavior.  
As shown in Table~\ref{tab:proficiency_ego_hrv_60s}, the our method improves the overall accuracy by a clear margin, reinforcing the need for higher temporal resolution in egocentric settings.

\subsection{Visual examples}
\begin{figure}[h]
    \centering
    \includegraphics[width=\linewidth]{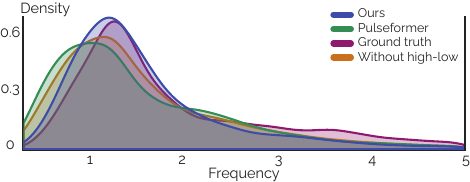}
    \caption{Spectral similarity between model outputs and ground-truth is evaluated using kernel density estimates. Our method shows the highest overlap with the true contact PPG frequency distribution, indicating better recovery of relevant frequency components.}
    \label{fig:append_freq}
\end{figure}

To further illustrate our approach, we visualize the frequency characteristics of the reconstructed BVP signals.  
Figure~\ref{fig:append_freq} shows the magnitude spectra obtained using the Fourier transform for the outputs of different models alongside the ground truth BVP from the contact nose PPG sensor.  
Our method exhibits the closest spectral distribution to the reference, demonstrating improved recovery of physiologically relevant frequency components.

\subsubsection{Spectral distribution analysis}
To quantitatively assess how closely each reconstructed BVP signal matches the spectral profile of the reference contact PPG, we evaluate the divergence between their normalized power spectral densities (PSDs).  
Let $P_i(f)$ denote the average PSD distribution of model $i$ over frequency $f \in \mathcal{F}$, normalized to form a probability density.
\begin{equation}
p_i(f) = \frac{P_i(f)}{\sum_{f \in \mathcal{F}} P_i(f)} , \quad
\sum_{f \in \mathcal{F}} p_i(f) = 1.
\end{equation}

For two distributions $p_i(f)$ and $p_j(f)$, the \textit{Kullback–Leibler divergence} (KL) is defined as in Equation~\ref{eq:KL}.
\begin{equation}\label{eq:KL}
D_{\mathrm{KL}}\!\left(p_i \,\|\, p_j\right)
= \sum_{f \in \mathcal{F}} p_i(f) \, \log \frac{p_i(f)}{p_j(f)} .
\end{equation}
KL divergence is asymmetric, i.e.\ $D_{\mathrm{KL}}(p_i \| p_j) \neq D_{\mathrm{KL}}(p_j \| p_i)$,  
where larger $D_{\mathrm{KL}}(p_i \| p_j)$ indicates that $p_i$ allocates energy in frequency regions absent in $p_j$.  
To obtain a symmetric and bounded measure, we also report the \textit{Jensen–Shannon divergence} (JS). 
We compute divergences between the predicted BVP distributions and the reference contact PPG ($p_{\mathrm{ref}}$).  
Table~\ref{tab:freq_js} summarizes the results.

\begin{table}[h]
\centering
\caption{Spectral divergence between normalized PSDs of model outputs and the reference contact PPG.  
Lower is better.}
\label{tab:freq_js}
\renewcommand{\arraystretch}{1.05}
\begin{adjustbox}{width=1\columnwidth,center}
\begin{tabular}{lccc}
\toprule
\textbf{Comparison} & $D_{\mathrm{KL}}(p_i\|p_{\mathrm{ref}})$ & $D_{\mathrm{KL}}(p_{\mathrm{ref}}\|p_i)$ & $D_{\mathrm{JS}}(p_i,p_{\mathrm{ref}})$ \\
\midrule
Ours vs. contact PPG           & 0.0492 & 0.0577 & \textbf{0.0131} \\
Without High–Low vs. contact PPG & 0.0919 & 0.0755 & 0.0198 \\
PulseFormer vs. contact PPG    & 0.1659 & 0.1256 & 0.0337 \\
\bottomrule
\end{tabular}
\end{adjustbox}
\end{table}

Our model achieves the lowest divergence from the ground-truth spectrum ($D_{\mathrm{JS}}{=}\,0.0131$),  
indicating the highest fidelity in recovering frequency content.  
Removing the low–high decomposition module increases $D_{\mathrm{JS}}$ by 51\%, while the baseline PulseFormer model diverges by 157\%.  
The asymmetry in $D_{\mathrm{KL}}$ suggests that competing models tend to overemphasize frequencies absent in the true PPG spectrum,  
whereas our approach yields a closer match across both low and high-frequency bands relevant to the pulse.

\begin{figure}[h]
    \centering
    \includegraphics[width=0.9\linewidth]{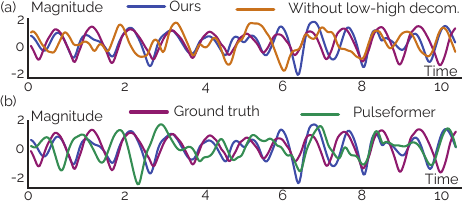}
    \caption{
    Qualitative comparison of reconstructed waveforms.
    (a) compares our method (blue) against the ground-truth contact BVP (purple) and the ablated variant without low–high decomposition (yellow).  
    (b) compares our method (blue) with PulseFormer~\cite{egoPPG} (green) for the same segment.
    Our approach aligns closely with the ground-truth waveform, whereas PulseFormer and the ablated model exhibit phase misalignment.
    }
    \label{fig:time_domain_ppg}
\end{figure}
\vspace{-5mm}
\subsubsection{Time-domain analysis}
To further assess temporal fidelity, we visualize reconstructed pulse waveforms for a representative 10-second sequence in Figure~\ref{fig:time_domain_ppg}.  
Our method maintains both phase and amplitude consistency with the ground-truth contact BVP signal, accurately capturing beat-to-beat variations.  
In contrast, PulseFormer~\cite{egoPPG} and the model without the low–high decomposition exhibit temporal lag and smoothed pulse contours, confirming that the proposed decomposition enhances precise recovery of cardiac cycles from egocentric video.

\subsection{Evaluation of uncertainty calibration}
\label{sec:eval_of_calibrate}
\begin{figure}[h]
    \centering
    \includegraphics[width=0.5\linewidth]{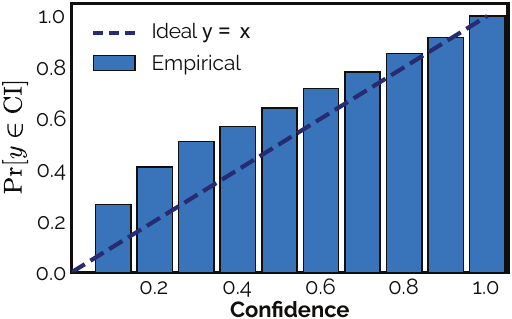}
    \caption{Reliability diagram of empirical versus nominal coverage. The model exhibits slightly conservative uncertainty at low confidence and near-perfect calibration at high confidence.}
    \label{fig:uncertainty}
\end{figure}

\begin{figure*}[h]
\centering
\begin{subfigure}[t]{0.48\textwidth}
  \vspace{-4mm} 
  \centering
  \includegraphics[width=\linewidth]{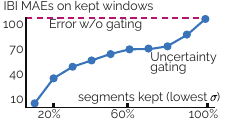}
  \caption{Coverage--error trade-off: mean IBI MAE vs. fraction retained (lowest uncertainty).}
  \label{fig:coverage}
\end{subfigure}
\hfill
\begin{subfigure}[t]{0.48\textwidth}
  \vspace{0pt}
  \centering
  \renewcommand{\arraystretch}{1.05}
  \setlength{\tabcolsep}{5pt}
  \begin{adjustbox}{width=\linewidth}
  \begin{tabular}{lccc}
    \toprule
      & \textbf{All} & \textbf{80\%} & \textbf{20\%} \\
    \midrule
    \textbf{RMSSD-MAE}  & 19.14 & 13.27 & \textbf{10.78} \\
    \textbf{RMSSD-MAPE} & 42.55 & 28.12 & \textbf{17.13} \\
    \midrule
    \textbf{IBI-MAE}    & 105.91 & 73.42 & \textbf{39.58} \\
    \textbf{IBI-MAPE}   & 13.56 & 9.87 & \textbf{6.74} \\
    \bottomrule
  \end{tabular}
  \end{adjustbox}
  \caption{Uncertainty gating reduces error (all vs lowest-uncertainty 80\% and 20\%).}
  \label{tab:gated_error}
\end{subfigure}
\vspace{-1mm}
\caption{Uncertainty correlates with error and supports reliability gating.}
\label{fig:uncertainty_gating_block}
\end{figure*}

We propagate the predictive variance using the delta method and then apply a single affine variance calibration.  
The reliability diagram (Fig.~\ref{fig:uncertainty}) shows empirical coverage versus nominal confidence.  
Uncertainty estimates are informative (Spearman’s $\rho$=0.714) and calibrated intervals are slightly conservative at low confidence (10\%\,$\to$\,30.9\%, 20\%\,$\to$\,46.7\%) but align well at high confidence (90\%\,$\to$\,91.6\%).  
We compute the normalized Area Under the Calibration Error Curve (AUCE=0.217) for the total deviation from perfect calibration~\cite{Scalia2020}.  
Results show low miscalibration, and strong agreement at high confidence.
We also compute the relationship between predicted uncertainty and error where higher uncertainty corresponds to higher error (Fig~\ref{fig:coverage}, coverage vs IBI error). 
Using uncertainty for gating segments improves overall estimation reliability.
In other words, keeping lower-uncertainty windows reduces errors.

Finally, treating high-error windows as failures ($|\mathrm{IBI\, error}|>50$\,ms), uncertainty predicts these segments with $\mathrm{AUROC}=0.715$, confirming that uncertainty is informative for trustworthy prediction. 

\subsection{Breakdown of errors in egoPPG-DB}
\label{sec:breakdown_of_errors}
We analyze error variation across activities.
Errors are lowest in low-motion segments (e.g., standing or light head movements) and increase in high-motion activities.
Table~\ref{tab:per_activity_4s_ours_vs_pf} reports the per-session results as reported in egoPPG-DB.
\begin{table*}[htbp]
\centering
\caption{Per-activity breakdown on egoPPG-DB at 4\,s resolution. We report our method and PulseFormer~\cite{egoPPG}.}
\label{tab:per_activity_4s_ours_vs_pf}
\begin{adjustbox}{width=1.0\columnwidth,center}
\renewcommand{\arraystretch}{1.1}
\begin{tabular}{lcccccccc}
\toprule
\multirow{2}{*}{\textbf{Activity}} 
& \multicolumn{2}{c}{\textbf{HR MAE}} 
& \multicolumn{2}{c}{\textbf{IBI MAE [ms]}} 
& \multicolumn{2}{c}{\textbf{SDNN MAE [ms]}} 
& \multicolumn{2}{c}{\textbf{RMSSD MAE [ms]}} \\
\cmidrule(lr){2-3}\cmidrule(lr){4-5}\cmidrule(lr){6-7}\cmidrule(lr){8-9}
& \textbf{PF} & \textbf{Ours}
& \textbf{PF} & \textbf{Ours}
& \textbf{PF} & \textbf{Ours}
& \textbf{PF} & \textbf{Ours} \\
\midrule
Kitchen  & 18.32 & \textbf{10.35} & 125.1 & \textbf{88.50}  & 19.13 & \textbf{8.90} & 25.17  & \textbf{16.53} \\
Office   & 16.93 & \textbf{ 8.83} & 115.0 & \textbf{80.73}  & 17.64 & \textbf{7.80} & 30.68  & \textbf{14.66} \\
Dancing  & 26.01 & \textbf{14.21} & 165.6 & \textbf{118.0}  & 41.30 & \textbf{12.80} & 60.15  & \textbf{23.87} \\
Bike     & 34.48 & \textbf{17.87} & 210.5 & \textbf{141.7}  & 55.19 & \textbf{16.90} & 90.68  & \textbf{31.20} \\
Walking  & 16.50 & \textbf{10.66} & 122.8 & \textbf{93.01}  & 33.75 & \textbf{10.10} & 50.23  & \textbf{19.63} \\
\midrule
\textbf{Overall} & 22.10 & \textbf{12.72} & 147.5 & \textbf{105.9} & 33.71 & \textbf{10.89} & 53.90 & \textbf{19.14} \\
\bottomrule
\end{tabular}
\end{adjustbox}
\end{table*}

Table~\ref{tab:per_activity_4s_ours_vs_pf} shows substantial variation across activities.
In low-motion sessions (e.g., office and kitchen), errors drop markedly, and the resulting HR/HRV estimates approach ranges commonly reported for contact-based estimation in non-clinical settings~\cite{Choi_2017, BS_HRV}.
During high-motion activities, errors increase, but our uncertainty estimates rise accordingly.
This enables identifying unreliable windows for down-weighting or exclusion to improve the trustworthiness of the remaining segments.
We also observe a direct downstream benefit.
When we remove the uncertainty from PhysFusion, the performance decreases for the proficiency classification. (Table~\ref{tab:ablation_proficiency}).

\clearpage
\subsection{Further results regarding the sampling rate}
\label{sec:egoexo4d_std}
We recorded an additional dataset with 10 subjects at 90\,fps to evaluate whether higher temporal resolution improves HR and HRV estimation.
We compare the same pipeline at 30\,fps and 90\,fps without changing the temporal length of the windows (4 or 60\,s).
Results in Table~\ref{tab:hr_hrv_results_fps} show that 90\,fps yields only marginal gains, while 30\,fps remains highly competitive for HR and sufficiently accurate for HRV indices. 
This supports using 30\,fps as a practical operating point with lower compute and storage cost, without sacrificing performance.

\begin{table*}[htbp]
\centering
\caption{Comparison of heart rate (HR) and heart rate variability (HRV) estimation at 30\,fps and 90\,fps. The 90\,fps setting yields slightly lower errors, while 30\,fps remains sufficient for HRV indices and performs strongly for HR.}
\label{tab:hr_hrv_results_fps}
\begin{adjustbox}{width=1\columnwidth,center}
\renewcommand{\arraystretch}{1.1}
\begin{tabular}{lcccccccccccccccccccc}
\toprule
\multirow{3}{*}{\textbf{Method}} 
& \multicolumn{8}{c}{\textbf{HR}} 
& \multicolumn{8}{c}{\textbf{HRV (IBI) [ms]}} 
& \multicolumn{4}{c}{\textbf{HRV Time-Domain [ms]}} \\
\cmidrule(lr){2-9}\cmidrule(lr){10-17}\cmidrule(lr){18-21}
& \multicolumn{2}{c}{\textbf{MAE}} & \multicolumn{2}{c}{\textbf{RMSE}} & \multicolumn{2}{c}{\textbf{MAPE (\%)}} & \multicolumn{2}{c}{\textbf{$r$}}
& \multicolumn{2}{c}{\textbf{MAE}} & \multicolumn{2}{c}{\textbf{RMSE}} & \multicolumn{2}{c}{\textbf{MAPE (\%)}} & \multicolumn{2}{c}{\textbf{$r$}}
& \multicolumn{2}{c}{\textbf{SDNN}} & \multicolumn{2}{c}{\textbf{RMSSD}} \\
\cmidrule(lr){2-3}\cmidrule(lr){4-5}\cmidrule(lr){6-7}\cmidrule(lr){8-9}
\cmidrule(lr){10-11}\cmidrule(lr){12-13}\cmidrule(lr){14-15}\cmidrule(lr){16-17}
\cmidrule(lr){18-19}\cmidrule(lr){20-21}
& \textbf{4\,s} & \textbf{60\,s} & \textbf{4\,s} & \textbf{60\,s} & \textbf{4\,s} & \textbf{60\,s} & \textbf{4\,s} & \textbf{60\,s}
& \textbf{4\,s} & \textbf{60\,s} & \textbf{4\,s} & \textbf{60\,s} & \textbf{4\,s} & \textbf{60\,s} & \textbf{4\,s} & \textbf{60\,s}
& \textbf{MAE} & \textbf{MAPE} & \textbf{MAE} & \textbf{MAPE} \\
\midrule
Ours (30\,fps) & 8.85 & 5.33 & 12.72 & 8.90 & 10.03 & 5.78 & 0.66 & 0.91
                        & 81.73 & 62.16 & 102.93 & 86.48 & 8.05 & 6.33 & 0.71 & 0.85
                        & 7.90 & 22.15 & 15.02 & 33.57 \\
Ours (90\,fps) & 8.72 & 5.12 & 12.45 & 8.58 & 9.41 & 5.41 & 0.68 & 0.92
              & 78.60 & 60.90 & 99.10 & 84.20 & 7.85 & 6.15 & 0.74 & 0.86
              & 7.55 & 21.10 & 14.20 & 30.10 \\
\bottomrule
\end{tabular}
\end{adjustbox}
\end{table*}
In this fps configuration, we observe error rates in the same range as those reported for EgoPPG in the office scenario, indicating that our measurements and evaluation protocol are consistent and that 30\,fps is sufficient for reliable HR and HRV estimation in this setting.


\section{Expanded discussion and related work}
\label{sec:expanded_discussion}

\paragraph{Why eye-video?}
Contact PPG requires consistent skin coupling and pressure~\cite{PPG_regulation}, which can vary with eyewear fit, slippage during motion, and user comfort over long wear~\cite{comfort_healthsense, comfort_2}.
These factors introduce practical failure modes and add mechanical overhead.
Since many egocentric platforms already include inward-facing cameras for interaction~\cite{noauthor_magic_nodate, engel2023projectarianewtool}, our goal is to leverage existing sensing to provide a \emph{no-additional-contact-hardware} option for physiological estimation.
We do not argue that eye-video should replace contact PPG when the sensors are available; rather, we target settings where adding or relying on contact sensing is impractical (long-term wear or studies prioritizing minimal instrumentation).

\paragraph{How can 30\,fps be feasible for HRV?}
HRV is timing-sensitive, so it is reasonable to question whether 30\,Hz eye video can support HRV estimation.
Our method estimates HRV from a BVP signal, where prior work has reported stable HRV estimates at low sampling rates (e.g., 25\,Hz) by computing HRV on an interpolated beat-to-beat representation~\cite{9495237}.
Empirically, downsampling BVP to 25\,Hz has been shown to change SDNN and RMSSD only marginally (on the order of a few milliseconds)~\cite{Choi_2017}, suggesting that sampling rate alone is not the dominant failure mode when the waveform is clean.

In our setting, the main challenge is not the frame rate but the corruption in the recovered BVP.
We address this by (i) improving BVP fidelity from 30\,Hz video with low-high decomposition, and (ii) refining temporal structure before HRV computation via interpolation in a learned frequency-domain module (U-Net), rather than relying on raw frame-level peak timing.
Finally, we model uncertainty.
Accordingly, we position our method as a \emph{non-clinical}, wearable HRV estimation for behavior modeling, with calibrated confidence to indicate when the estimates should be trusted.

\paragraph{Are 4\,s estimates sufficient?}
Our approach predicts the mean IBI over non-overlapping 4\,s windows, whereas classical HRV pipelines compute IBIs from individual R-peaks.
In our setting, peak-to-peak IBIs are sensitive to noise and missed/spurious peaks, which can dominate the error.
To assess whether 4\,s windowing itself introduces a limiting bias, we ran an oracle analysis on the ECG.
We computed window-level mean IBIs from the ground-truth ECG using the same 4\,s windows and compared them to beat-level IBIs.
The resulting discrepancy was only 5--6\,ms, which is substantially smaller than the current prediction error. This suggests that the 4\,s aggregation is not the main bottleneck.

\paragraph{Role of IMU}
IMU is used as an auxiliary motion context cue, not as a physiological sensor.
Its placement is fixed in the device coordinate system.
We therefore avoid relying on absolute IMU scale and treat IMU features as weak conditioning.
Our method limits sensitivity to IMU orientation by using its magnitude while still benefiting from motion cues when available on the egocentric platform.



\paragraph{Generalization and dataset limitations.}
Our evaluation is constrained by publicly available datasets~\cite{egoPPG} with synchronized ground truth and egocentric eye video.
This limits coverage of conditions such as outdoor illumination and extreme motions.
We view this as a key next step where broader data collection and condition-wise analysis (motion intensity, lighting proxies, speaking activity) can better characterize operating regimes and guide more robust modeling.





%
%


\end{document}